# Calamari – A High-Performance Tensorflow-based Deep Learning Package for Optical Character Recognition

Christoph Wick, Christian Reul, and Frank Puppe
Universität Würzburg, Chair of Computer Science VI
{prename.surname}@uni-wuerzburg.de

**Abstract**

*Optical Character Recognition (OCR) on contemporary and historical data is still in the focus of many researchers. Especially historical prints require book specific trained OCR models to achieve applicable results (Springmann and Lüdeling, 2016, Reul et al., 2017a). To reduce the human effort for manually annotating ground truth (GT) various techniques such as voting and pretraining have shown to be very efficient (Reul et al., 2018a, Reul et al., 2018b). Calamari is a new open source OCR line recognition software that both uses state-of-the art Deep Neural Networks (DNNs) implemented in Tensorflow and giving native support for techniques such as pretraining and voting. The customizable network architectures constructed of Convolutional Neural Networks (CNNS) and Long-Short-Term-Memory (LSTM) layers are trained by the so-called Connectionist Temporal Classification (CTC) algorithm of Graves et al. (2006). Optional usage of a GPU drastically reduces the computation times for both training and prediction. We use two different datasets to compare the performance of Calamari to OCRopy, OCRopus3, and Tesseract 4. Calamari reaches a Character Error Rate (CER) of 0.11% on the UW3 dataset written in modern English and 0.18% on the DTA19 dataset written in German Fraktur, which considerably outperforms the results of the existing softwares.*

## 1 Introduction

Optical Character Recognition (OCR) of contemporary printed fonts is widely considered as a solved problem for which many commercial and open source software products exist. However, the task of text recognition on early printed books is still a challenging task due to a high variability in typeset, additional characters, or low scan quality. To master OCR on early printed books, a book or font specific model must be trained to achieve CERs below 2% (Reul et al., 2017a, Springmann and Lüdeling, 2016). For this purpose, highly performant individual models must be trained in a short period of time using as less manually annotated GT files as possible. Currently, there exist several Free Open Source Software (FOSS) attempts for such programs: OCRopy, OCRopus 3, Kraken, or Tesseract 4, each with its own advantages or drawbacks.

Calamari[1] is a novel software for training and applying OCR models on text lines including several up-to-date techniques to highly optimize the computation time and the performance of the models. The usage of Tensorflow allows to design arbitrary DNNs including CNN and LSTM structures that are proven to yield improved results compared to shallow network structures (Breuel, 2017, Wick et al., 2018). These networks can optionally use CUDA and cuDNN (on a supported GPU) which results in a highly reduced computation time. Compared to other FOSS Calamari supports various techniques that minimize the CER including voting and pretraining (see Reul et al., 2018a, Wick et al., 2018). The software is not designed as a full OCR pipeline which includes tasks such as layout analysis, or line segmentation, but is the topic of a separate publication (in preparation), instead it focuses solely on the OCR step that transcribes line images to text. However, Calamari as Python-Package[2] can easily be integrated in existing pipelines to manage the OCR part. Thus, by design without any changes it can directly replace the OCR-Engine of OCRopy.

---

[1] https://github.com/Calamari-OCR
[2] https://pypi.org/project/calamari_ocr/

## 2   Related Work

In the following, we give a short list of the existing open source OCR programs OCRopy, OCRopus 3, Tesseract 4, and Kraken. All of these systems are designed to support the full pipeline from plain page to text. Since the intention of Calamari is solely to handle the OCR of line images, only this functionality of the other programs will be described and compared.

Currently, there exist several versions of OCRopy which was originally published by Breuel (2008) (see also Breuel et al., 2013), that are still maintained. OCRopy[3] is the first software that allowed a user to train custom LSTM based models incorporating the CTC-Loss function (Graves et al., 2006). By default, it uses a slow numpy based implementation of the computation, which can be exchanged by a faster C-based clstm one. However, neither the GPU nor Deep CNN-LSTM models can be used. For training it requires a list of images and their GT and outputs a model, which then can be used to predict the written text of other text lines.

Kraken[4] is a fork of OCRopy, which has a different API, uses clstm as default, and adds support for bidirectional and top to bottom text. Currently, the usage of PyTorch[5] as Deep Learning engine supporting GPU training is under development.

While OCRopy is still maintained OCRopy 2[6] seems not to be developed anymore, probably due to the introduction of OCRopus 3[7] which changed all major OCR components to Deep Models using PyTorch. OCRopy 3 supports variable network architectures including CNNs and LSTMs and allows training and applying of the models on the GPU. The resulting models yield state-of-the-art results and can be trained with minimal effort in time.

Tesseract[8] was initially released as open source in 2005 and is still under development. The newest version 4 of Tesseract adds support for deep network architectures such as LSTM-CNN-Hybrids, however GPU support is not offered. To prototype network structures Tesseract proposes a Variable-size Graph Specification Language (VGSL) which is similar to the network prototype language of Calamari.

## 3   Methods

Calamari comprises several techniques to achieve state-of-the art results on both contemporary prints and historical prints. Beside different DNN architectures it supports confidence voting of different predictions and finetuning with codec adaption. These methods will be presented in the following.

### 3.1   Network Architecture Building Blocks

The task of the DNN and its decoder is to process the image of a segmented text line and to output simple text. This sequence to sequence task is trained by the CTC algorithm published by Graves et al. (2006) that allows to predict shorter but arbitrary label sequences out of an input sequence that is the line image regarded as sequence. Hereby, the network outputs a probability distribution for each possible character in the alphabet for each horizontal pixel position of the line. Thus, an image with size $h \times w$ with a given alphabet size of $|L|$ will result in a matrix of shape $P(x,l) \in \mathbb{R}^{w \times |L|}$ with $P(x,l)$ being a probability distribution for all $x$. Since the network needs to see several slices in width to be certain about a single character, most of the time it does not yet know what to predict. Hence, the CTC-algorithm adds a *blank* label to the alphabet that is ignored by the decoder but allows the network to make empty or uncertain predictions with a high probability. In fact, the used greedy decoder that chooses the character with the highest probability at each position $x$ mostly predicts blank labels, and only with one or two pixel widths the actual character is recognized. Afterwards, the final decoding result is received by unifying neighbouring predictions of the same characters and removing all blank labels. For example the sentence $AA--B--CA--A-$ of length w is reduced to $ABCAA$.

---

[3] https://github.com/tmbdev/ocropy
[4] http://kraken.re/
[5] https://pytorch.org/
[6] https://github.com/tmbdev/ocropy2
[7] https://github.com/NVlabs/ocropus3
[8] https://github.com/tesseract-ocr/tesseract

The network is trained by the CTC-Loss-Function that basically computes the probability of the GT label sequence given the probability output of the network. This probability is computed by summing up all possible paths through the probability matrix P that yield the GT using an efficient forward backward algorithm. Fortunately, this computation can be derived to receive the gradient that is required for the learning algorithm.

The supported network architectures of Calamari are CNN-LSTM-Hybrids that act on a full line in one step. CNNs are widely used in image processing because they are designed to detect meaningful features (e.g. curves, circles, lines, or corners) that can be located anywhere in an image. These features are processed afterwards by a (bidirectional) LSTM based recurrent network to compute the probability matrix $P$. Max-Pooling is a common technique in CNNs to reduce the computation effort and keep only the most important features. In general image processing settings pooling is applied both in vertical and horizontal direction. This means that the processed lines get shortened. Thus, it is important that the final layer of the CNN in the full network has an image width, that is long enough to produce the full label sequence. For example, if the GT has a length of 40 characters, a minimum of 80 predictions are required to allow for a blank prediction between any pair of adjacent characters. Thus, if two $2 \times 2$ pooling layers are used in the CNN the width of the image lines should be at least $w = 2 \cdot 2 \cdot 80\text{px} = 320\text{px}$.

## 3.2 Finetuning and Codec Resizing

A general approach to improve the accuracy of a model on a specific dataset is not to train from scratch but instead to start from an existing robust model and to finetune for the specific task (see e.g. Reul et al., 2017b). Usually, the alphabet of the base model differs from the desired labels which is why the output layer is usually fully replaced. However, in the task of OCR many letters, digits, or punctuations are shared across the base model and the target task, and only a few new letters might be added e.g. German umlauts when starting from an English model, or erased e.g. the character "@" which does not exist in historical texts. It is rational to keep those weights as is and add or remove new or unneeded labels instead of training the output layer from scratch.

In the area of historical printed books an individual model for each book must be trained to achieve reasonable results for OCR. To reduce the human effort required for manually transcribing GT lines the OCR model should be trained using as few lines as possible. However, if using only a few lines some characters might not be present yet, e.g. capital letters or digits. Hence, a whitelist is useful to define characters that should not be erased from the base model. Thus, the resulting model has still a chance to predict those letters, even if they have never been seen during finetuning.

$$\text{An examp} \left\{ \begin{array}{l|l|l} \text{l} \quad \textbf{0.8\%} & \text{I} \quad 0.2\% & \text{L} \quad 0.0\% \\ \text{l} \quad 0.4\% & \textbf{I} \quad \textbf{0.5\%} & \text{L} \quad 0.1\% \\ \text{l} \quad 0.2\% & \textbf{I} \quad \textbf{0.3\%} & \text{L} \quad 0.2\% \end{array} \right\} e$$

*Figure 1: An example for the confidence voting algorithm. Each row shows a part of the output of three different voters. When choosing the most frequent top result of each voter (bold) an "I" would be predicted. However, when adding the confidences of each voter, the letter "l" is predicted.*

Another benefit of using a pretrained model is the reduced computation time to train a model. Since the initial weights are not randomly chosen but set to meaningful features that are expected to generalize on the new data only small variations are required to optimize on the new data.

### 3.3 Voting

Another technique to obtain more robust results is to vote the outcomes of different models (see e.g. Reul et al., 2018b). The benefit of voting depends highly on the variance of the individual voters. If the voters predict very similar results, errors are less probable of being removed, as if more diverse models are used.

In case of OCR, confidence voting showed the best results so far. This voting mechanism does not only include the most likely predicted character but also alternatives with its probabilities into account. Figure 1 shows a simple example of confidence voting. Three different voters predict the possible characters with an individual confidence. If a single voter chooses the character with the highest confidence, the letter "I" is winning in a majority vote. However, if one adds up each individual confidence values the correct letter "l" is chosen as the final output.

To obtain different voters based on a dataset several approaches are meaningful, e.g. using different training data, network architectures, base models for finetuning. Recently, we showed that variable voters can be generated by a simple but robust approach called cross-fold-training (Reul et al., 2018b).

## 4 The Calamari OCR-System

Calamari supports easy instructions to use any of the listed methods to train various models, and to apply them on existing lines. The software is implemented in Python3 and uses Tensorflow for Deep Learning of the neural net. In doing so, Calamari supports usage of the GPU including the highly optimized cuDNN kernels provided by NVIDIA for a rapid training and prediction of multiple lines (batches) simultaneously.

### 4.1 Preprocessing

Both the lines and the text are preprocessed by Calamari for all conducted experiments. The line images are converted to grayscale and are rescaled proportionally to a standard height of 48 pixels. Optionally, the lines are dewarped using OCRopy's dewarping algorithm. Crucial problems of the bidirectional LSTM are the predictions of the first and last few pixels of a line which can be seen as transient behaviour of the internal LSTM state. Therefore, a padding of 16 white pixels is added to each side of the line.

The textual preprocessing allows to resolve several visual ambiguities such as converting Roman unicode digits to Latin letters or joining repeated white space. Furthermore, Calamari adds support for mixed left-to-right and right-to-left text. This solves a challenging task: mirrored symbols e.g. opening or closing brackets depend on the reading order which can change within a line.

## 4.2 Training

The default network consists of two pairs of convolution and pooling layers with a ReLU-Activation function, a following bidirectional LSTM layer, and an output layer which predicts probabilities for the alphabet. Both convolution layers have a kernel size of 3 × 3 with zero padding of one pixel. The first layer has 64 filters, the second layer 128 filters. The pooling layers implement Max-Pooling with a kernel size and stride of 2 × 2. Each LSTM layer (forwards and backwards) has 200 hidden states that are concatenated to serve as input for the final output layer. During training we apply dropout (Srivastava et al., 2014) with a rate of 0.5 to the concatenated LSTM output to prevent overfitting. The loss is computed by the CTC-Algorithm given the output layer's predictions and the GT label sequence.

Calamari uses Adam (Kingma and Ba, 2014) as standard solver with a learning rate of 0.001. To tackle the exploding gradient problem of the LSTMs we implement gradient clipping on the global norm of all gradients as recommended by Pascanu et al. (2013).

As input for training Calamari expects just as OCRopy a list of line images and the corresponding text files. For efficiency, the full data is loaded and kept in memory for the complete training task instead of repeatedly reading only the current training example.

## 4.3 Prediction

To apply a trained model on new line images Calamari expects one or more models for prediction. If several models are used, Calamari votes the results of each individual model to generate the output sentence.

Sometimes, it might be useful to access additional information of the prediction. Therefore, Calamari allows to generate information about the position and its confidence of each individual character, as well as the full probability distribution of the input.

# 5 Experiments

To compare the performance of Calamari to OCRopy, OCRopus 3, and Tesseract 4 we use the datasets UW3 and DTA19. All final results of Calamari were achieved by using early stopping. Hereby we check every 20,000 iterations for a smaller CER on 20% of the training data (hence 80% are used for actual training), after ten checks without a better model, we interrupt the training. Similarly, we chose the best OCRopy or OCRopus 3 model based on the same 20% of training data. Tesseract 4 itself chooses its best model on the provided test data set. For Calamari and OCRopus 3 we use a batch size of 5, the OCRopy and Tesseract 4 do not support batching.

## 5.1 Datasets

For evaluating Calamari, we use two datasets with several millions of characters in the training set and three historical printed books.

The smaller *University of Washington Database III*[9] (UW3) consists of pages of modern English prints with more than 70,000 lines for training and almost 10,000 lines for evaluation. Breuel (2017) uses a different version of this dataset (UW3α) with 95,190 text lines for training and 1,253 text lines for evaluation. Unfortunately, this split is not publicly available.

The other large dataset *German Text Archive of the 19th Century*[10] (DTA19) extracted from the German Text Archive (see Wiegand et al., 2018) consists of 39 German Fraktur novels printed during the 19th century. Eight novels consisting of 50,968 text lines are used for evaluation the remaining books with 192,974 lines for training. Thus, the evaluation measures the capability of the models to generalize for unseen fonts.

---

[9] http://www.tmbdev.net/ocrdata-split/
[10] To be published

Figure 2: Example lines of the used UW3, DTA19, 1476, 1478, and 1499 datasets. Note that the last line is cropped.

Table 1: Overview of the used datasets. The Codec size lists the number of possible characters in the alphabet. The GT lines, the number of total characters, and the average line width in pixels are both shown for the training and the evaluation data sets. The average width of the evaluation set of UW3α is unknown.

| ID | Lang. | Code | Training GT Lines | Training Chars | Training Width | Evaluation GT Lines | Evaluation Chars | Evaluation Width |
|---|---|---|---|---|---|---|---|---|
| UW3 | English | 87 | 72,807 | 3,493,308 | 973 | 9,729 | 469,171 | 977 |
| UW3α | English | 87 | 95,190 | 4.5 Mio | 854 | 1,253 | 60 K | – |
| DTA19 | German | 159 | 192,974 | 8,711,800 | 912 | 50,968 | 2,304,242 | 905 |
| 1476 | German | 64 | 50 | 1,528 | 489 | 3,110 | 102,137 | 508 |
| 1478 | German | 69 | 50 | 2,054 | 731 | 2,695 | 118,437 | 755 |
| 1499 | German | 73 | 50 | 4,139 | 1117 | 1,578 | 126,445 | 1031 |

The three historical printed books[11] of the years 1476, 1478, and 1499 are written in broken script (Fraktur in a wider sense) and only consist of 3,100, 2,695, and 1,578 lines respectively. We use only 50 lines for training to simulate a human annotator that has to manually transcribe GT data, yet wants to achieve the best possible results.

An example line for each dataset is shown in Figure 2 and an overview of the datasets and their statistics are shown in Table 1. The codec size lists the number of characters in the alphabet. The average line width is computed after preprocessing (see Sec. 4.1) which forces a line height of 48 pixels

## 5.2 Evaluation Measure

To measure the performance of each OCR system we use the CER which is defined as the edit distance (ed) of two sequences $s_1$ and $s_2$ normalized by the maximum length:

$$\text{CER} = \frac{\text{ed}(s_1, s_2)}{\max(|s_1|, |s_2|)}$$

This measure is 0 if all characters and 1 if no character match.

## 5.3 Accuracy

Table 2 lists all results for the different datasets, software and used network architectures. First, it is notable that OCRopy yields the worst results both on the UW3 and the DTA19 dataset due to the shallow network structure of only one hidden LSTM-layer.

---

[11] To be published

*Table 2: CER using different software on the UW3 and the DTA19 dataset. The convolutions C each have a kernel size of 3 × 3 with zero padding and a filter count of 64 in the first and 128 in the second layer. Max-Pooling Mp is used either with a kernel size and stride of 2 × 2 or 1 × 2. Note that when using 2 × 2 both the height and length of the line is halved while when used 1 × 2 only the height is halved. The last hidden layer is always a LSTM with 200 nodes. Calamari results are the average of five models that are then used to apply confidence voting. The CERs on UW3α were published by Breuel (2017).*

| Model | Voted | Software | Dataset | CER |
|---|---|---|---|---|
| C, Mp(2x2), C, Mp(2x2), LSTM(200) | No | Calamari | UW3 | 0.155% |
| C, Mp(2x2), C, Mp(2x2), LSTM(200) | Yes | Calamari | UW3 | 0.114% |
| LSTM(200) | No | OCRopy | UW3 | 0.870% |
| C, Mp(2x2), C, Mp(2x2), LSTM(200) | No | Tesseract 4 | UW3 | 0.397% |
| C, Mp(2x2), C, Mp(2x2), LSTM(200) | No | OCRopus3 | UW3 | 0.502% |
| C, Mp(1x2), C, Mp(1x2), LSTM(200) | No | OCRopus3 | UW3 | 0.436% |
| C, Mp(1x2), C, Mp(1x2), LSTM(100) | No | OCropus3 | UW3α | 0.25% |
| C, Mp(1x2), C, Mp(1x2), C, Mp(1x2), LSTM(200) | No | OCropus3 | UW3α | 0.25% |
| C, Mp(2x2), C, Mp(2x2), LSTM(200) | No | Calamari | DTA19 | 0.221% |
| C, Mp(2x2), C, Mp(2x2), LSTM(200) | Yes | Calamari | DTA19 | 0.184% |
| LSTM(200) | No | OCRopy | DTA19 | 1.59% |
| C, Mp(1x2), C, Mp(1x2), LSTM(200) | No | OCropus3 | DTA19 | 0.907% |

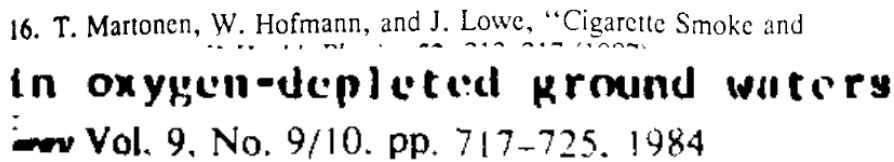

*Figure 3: Three of the worst recognized lines during the evaluation of Calamari using voting on the UW3 dataset. The upper two lines suffer from impurities of segmentation or font. The text of the bottom line is cropped.*

This network is not capable of learning and generalizing the variations of fonts in the datasets. Introduction of Convolution- and Pooling-Layers highly increases the accuracy of all software and on both datasets. Hereby, the same network structure *C, Mp(2x2), C, Mp(2x2), LSTM(200)* yields different results on the various frameworks: 0.155% on Calamari, 0.397% on Tesseract 4 and 0.502% on OCRopus3 for the UW3 dataset. This difference must be caused by a different training setup or varying hyperparameters, e.g. using an Adam-Solver or a Momentum-Solver, differences in the learning-rate, or usage of dropout. However, the evaluation of hyperparameters is out of the scope of this paper. Note, that the overall setup is the same: All frameworks use the same dataset split for training, choosing the best trained model and evaluation.

The CER on UW3α published by Breuel (2017) that is evaluated on a different evaluation split are in the same order of magnitude as the computed CER on the UW3 split used in this paper. Our trained model on OCRopus3 is comparable to the Tesseract 4 model.

On both models the best performance is achieved by Calamari and further improved by the voting mechanism. On the UW3 dataset it achieves impressive error rates of 0.11% without a language model such as a dictionary. Calamari with voting yields 0.18% CER on German Fraktur. This task is far more difficult because this dataset has a larger alphabet with very similar characters (e.g. a, à, á, â, or ä) or different fonts especially for capital letters. Yet, most errors are due to corrupt lines or GT inconsistencies (e.g. upper and lower quotation marks).

The evaluation on the UW3 dataset shows that 97% of all lines are recognized without any error, however the worst 0.1% lines contribute to 20% of the remaining error. Those lines are either wrongly segmented, rotated or highly degraded. Figure 3 shows three of the worst recognized examples. The first impurely segmented line contains the upper part of the following line. Therefore, the postprocessing step failed by bending the middle of the line and the line could not recognized anymore.

The second example has a degraded font, that is also larger than the other lines. For example, "waters" is misclassified as "wators" because the middle horizontal line in the "e" is missing. This error could surely be fixed by a dictionary.

The last line is falsely cropped at the beginning, thus it is impossible to recognize "Energy Vol. 9". Any other character is correctly recognized.

Table 3 lists the 20 most common errors of Calamari with and without voting on the UW3 dataset. The GT and PRED columns list the GT expectation and the prediction. The third column counts the number of errors in the test data set (≈0.47 million characters in total) and the last column the relative percentage to all errors made by the respective model. The last row adds the relative percentage of all errors that are not among the 20 most common. Note, that even though the missing '' occurred four times in the voted model the relative percentage is doubled, because the double quote is interpreted as two single quotes.

As expected the most common errors in both models are confusions of similar characters such as ".", and ",", "v" and "y", or "I" and "l", but also missing or inserted spaces. Most significantly, the voting mechanism reduces the error of missing spaces (25 vs 9 occurrences), but also the number of confusions of "e" and "c" dropped (7 to 3).

Table 3: Most common errors of Calamari on the UW3 evaluation set. The left side shows the errors of a single voter, the right side after voting. The first two columns show the GT and the predicted character, the third column the number of occurrences, and the forth column the relative percentage compared to the total CER, respectively. The last row sums the relative error remaining mistakes. Note that '' is interpreted as two single quotes, which is why the relative error is doubled if those two characters are missing.

| Single model | | | | Conf. Voted model | | | |
|---|---|---|---|---|---|---|---|
| GT | PRED | CNT | PCT | GT | PRED | CNT | PCT |
| , | . | 35 | 5.12% | , | . | 35 | 6.68% |
| ␣ |   | 35 | 5.12% | ␣ |   | 33 | 6.30% |
|   | ␣ | 25 | 3.66% | ' |   | 10 | 1.91% |
| . | , | 20 | 2.93% |   | ␣ | 9 | 1.72% |
| ' |   | 11 | 1.61% | . | , | 9 | 1.72% |
|   | , | 10 | 1.46% | y | v | 7 | 1.34% |
| . |   | 9 | 1.32% | , |   | 7 | 1.34% |
| i | l | 8 | 1.17% | i | l | 7 | 1.34% |
| y | v | 7 | 1.02% | . |   | 6 | 1.15% |
| e | c | 7 | 1.02% | I | l | 6 | 1.15% |
| n |   | 6 | 0.88% | e |   | 4 | 0.76% |
|   | i | 6 | 0.88% | '' |   | 4 | 1.53% |
| s |   | 5 | 0.73% |   | . | 4 | 0.76% |
| i | I | 4 | 0.73% | r |   | 3 | 0.57% |
| r |   | 4 | 0.59% | n |   | 3 | 0.57% |
|   | . | 4 | 0.59% | - |   | 3 | 0.57% |
| - | . | 4 | 0.59% | e | c | 3 | 0.57% |
| e |   | 4 | 0.59% |   | e | 3 | 0.57% |
|   | e | 4 | 0.59% | i |   | 3 | 0.57% |
| l | i | 4 | 0.59% | s |   | 3 | 0.57% |
|  | Remaining | 68.81% |  |  | Remaining | 68.32% |

Table 4: Average time for training or prediction of a single line of the UW3 dataset. Note that the times measured for OCRopy and Tesseract 4 are on the CPU while Calamari and OCRopy3 run on the GPU. The prediction of OCRopy and Tesseract 4 is evaluated using a single process, using multiple multithreading highly reduces their computation time. The last row was published by Breuel (2017).

| Model | Software | Training | Prediction |
|---|---|---|---|
| C, Mp(2x2), C, Mp(2x2), LSTM(200) | Calamari | 8 ms | 3 ms |
| LSTM(200) | OCRopy | 850 ms | 330 ms |
| C, Mp(2x2), C, Mp(2x2), LSTM(200) | Tesseract 4 | 1200 ms | 550 ms |
| C, Mp(2x2), C, Mp(2x2), LSTM(200) | OCRopy3 | 10 ms | 7 ms |
| C, Mp(1x2), C, Mp(1x2), LSTM(100) | OCRopy3 | – | 10 ms |

A common postprocessing step on OCR is to apply a dictionary on the recognized words. It is to be expected that errors on letters are highly reduced but punctuation errors are not decreased. Even in this case the voting mechanism is very useful since is reduces also whitespace and punctuation errors. However, note that in the field of OCR on historical pages a dictionary might not be present or even not desired if differences in spelling are of interest. In this field of research, the voting mechanism of Calamari is very useful to reduce the CER.

## 5.4 Recognition Speed

Another crucial measure besides its accuracy for a good OCR system is its speed. The runtimes of all softwares for training and prediction excluding preprocessing of a single line is listed in Table 4. The processing time of a single line highly depends on the network architecture, the line width, and the used hardware. All the time experiments were measured on a NVIDIA GTX 1080Ti GPU if the software supports GPU usage and an Intel Xeon E5-2690 CPU. Preprocessing requires ca. 50 ms per line with a single process and drops to 6 ms when processing eight lines in parallel.

Obviously, both OCRopus 3 and Calamari are faster than Tesseract or OCRopy by a factor of almost 100 since they support batched GPU training and prediction. Incorporation of a GPU therefore allows to recognize more than 100 lines per second.

The time for voting scales linearly with the number of models used as voters. Thus, the prediction time per line in the voting experiments is approximately five times higher than the results shown in Table 4. The time required for the aggregating the voting can be neglected.

## 5.5 Finetuning

Using the models trained on the UW3 or the DTA19 dataset to directly predict the books 1476, 1488, and 1499 yields discardable results with up to 50% CER. Thus, it is mandatory to train individual models requiring manually labelled GT data. In the following, we use 50 lines of the three historical printed datasets for training different models with and without using the pretrained models of UW3 and DTA19 from section 5.3. This amount of lines is very low, but shows that even a small amount of GT can drastically decrease the CER even though the trained model might not have seen all possible characters of the alphabet yet (e.g. capital letters). The CERs are shown in Table 5, whereby both the average of the five folds and their voting results are shown.

The 1476 and 1478 datasets behave similar. Both sets yield about 10% error when using an individual model and a lower CER when voting five different models. Using the pretrained UW3 and DTA19 models instead of training from scratch both the individual model CER and the voted CER drop significantly. Hereby, using DTA19 as pretrained model results in better models because the font and the German alphabet of DTA19 is closer to the historical prints than the modern English fonts of UW3.

Table 5: Results on the historical printed books using Calamari trained with 50 lines of GT and a batch size of 5. Both the results of using pretraining or training from scratch are indicated by second column. The CER lists the average of the five trained folds and the last column the voted result as CER.

| Dataset | Pretraining | CER | Voted |
|---|---|---|---|
| 1476 | – | 9.1% | 7.1% |
| 1476 | UW3 | 7.4% | 5.5% |
| 1476 | DTA19 | 5.5% | 4.0% |
| 1478 | – | 10.8% | 9.3% |
| 1478 | UW3 | 8.8% | 7.3% |
| 1478 | DTA19 | 8.6% | 6.6% |
| 1499 | – | 7.5% | 6.4% |
| 1499 | UW3 | 8.4% | 6.5% |
| 1499 | DTA19 | 6.6% | 4.7% |

Interestingly, pretraining on UW3 yields worse results on the 1499 dataset compared to the models without pretraining, but the voted results are similar. However, the model using DTA19 as initial values clearly predicts better values than the default model.

Reul et al. (2018a) showed that using the same pretrained model to train different voters yields worse voted results than using different pretraining models. Thus, it is to be expected that the overall results get further improved if one mixes the pretrained models of UW3 and DTA19. Of course, increasing the number of available lines of GT for training the respective book significantly improves the accuracies of the models, however this results in a higher amount of human effort to annotate GT.

# 6 Discussion

The conducted experiments clearly demonstrate the capabilities of Calamari on both contemporary and historical OCR tasks. Not only does Calamari yield outstanding performances compared to other OpenSource OCR software but also requires a minimum of time for training and prediction due to the usage of Tensorflow as Deep Learning framework including cuDNN. As already shown by Breuel (2017) and Wick et al. (2018), Deep Hybrid Convolutional-LSTM architectures increase the accuracies both on contemporary and historical prints. Our results have revealed significant differences of Tesseract 4, Ocropus 3 and Calamari due to variations of the network architectures and different sets of hyperparameters. It is to be expected that an optimization of these hyperparameters might further improve the accuracies of the OCR models. However, a very high amount of the remaining errors on UW3 and DTA19 are GT inconsistencies or impure lines, which are nearly impossible to predict correctly.

Voting and pretraining are two important mechanisms to increase the performance of newly trained models, especially if only few data is available. Voting has shown improved results on all conducted experiments, however on the cost of a higher prediction and training time, since several different models are independently used. Pretraining is most useful if the original model is similar to the new data and reduced both the CER and the training time. Especially, when training new individual models as required for OCR on historical prints pretraining should be used to receive the best possible results if only a few manually annotated lines are available.

# 7 Future Work

The main focus of this paper has been the presentation of Calamari as new line based OCR engine to replace OCRopy or Tesseract due to its low CER and fast computation times.

Although Calamari supports many features such as voting, or pretraining, plans for extensions exist. A first major work is data augmentation during training which is expected to significantly drop the CER especially if only a few lines are present. The augmentations basically are different types of noise, degradation, or generated background. Obviously, synthetic data based on existing fonts can also be incorporated for data augmentation.

Tesseract's language to define network topologies (VGSL) has a very simple and compact syntax. The current syntax of Calamari should also support this language to define networks.

Finally, since Calamari is designed very modular, it shall be extended to support other sequence-to-sequence tasks, such as monophonic Optical Music Recognition (e.g. Gregorian chants) or Speech-To-Text. All these tasks fundamentally share the tasks of processing two dimensional sequential data and output a sequence of classes. Only the data preprocessing e.g. of audio files and postprocessing is different. Calamari is designed to easily exchange these steps but keeping a generic structure for training, evaluation, and application.